\documentclass[10pt,twocolumn,twoside]{IEEEtran}
\usepackage{graphicx}
\usepackage{amsmath}
\usepackage{amssymb}
\usepackage[tight]{subfigure}
\usepackage{cite}
\usepackage{bm}
\usepackage{multirow}
\usepackage{booktabs}
\usepackage{makecell}
\usepackage{algpseudocode}
\usepackage{color}
\usepackage{arydshln}
\usepackage[dvipsnames]{xcolor}
\usepackage[export]{adjustbox}
\usepackage{enumerate}

\usepackage[pagebackref=true,breaklinks=true,colorlinks,bookmarks=false,linkcolor=blue,citecolor=blue]{hyperref}
\newcommand{\ublue}[1]{{\textcolor{blue}{#1}}}
\newcommand{\redb}[1]{\textcolor{red}{\textbf{#1}}}
\DeclareMathOperator*{\argmin}{arg\,min}
\begin{document}

\title{Incorporating Uncertainty-Guided and Top-k Codebook Matching for Real-World Blind Image Super-Resolution}

\author{Weilei~Wen, Tianyi~Zhang, Qianqian~Zhao, Zhaohui~Zheng, Chunle~Guo, Xiuli~Shao, and Chongyi~Li
	\thanks{W. Wen, T. Zhang, Q. Zhao, Z, Zheng, C. Guo, X. Shao, and C. Li are with the VCIP, College of Computer Science, Nankai University, Tianjin, 300350, China (e-mail: wlwen@mail.nankai.edu.cn). 
 
%  W. Ren is with the School of Cyber Science and Technology, Shenzhen Campus, Sun Yat-sen University, Shenzhen 518107, P.R. China (e-mail: rwq.renwenqi@gmail.com). 
 
%  H. Wang is with the College of Artificial Intelligence, Nankai University, Tianjin 300350, China, and also with the Shenzhen Research Institute of Nankai University, Shenzhen 518081, China (e-mail: hpwang@nankai.edu.cn).
     T. Zhang and Q. Zhao contributed equally to this work.

     Corresponding author: C. Guo (e-mail: guochunle@nankai.edu.cn)

}
}

\markboth{IEEE Transactions on Image Processing}{}

\maketitle

\begin{abstract}
Recent advancements in codebook-based real image super-resolution (SR) have shown promising results in real-world applications. The core idea involves matching high-quality image features from a codebook based on low-resolution (LR) image features. 
However, existing methods face two major challenges: inaccurate feature matching with the codebook and poor texture detail reconstruction. To address these issues,
we propose a novel Uncertainty-Guided and Top-k Codebook Matching SR (UGTSR) framework, which incorporates three key components: (1) an uncertainty learning mechanism that guides the model to focus on texture-rich regions, (2) a Top-k feature matching strategy that enhances feature matching accuracy by fusing multiple candidate features, and (3) an Align-Attention module that enhances the alignment of information between LR and HR features. Experimental results demonstrate significant improvements in texture realism and reconstruction fidelity compared to existing methods. We will release the code upon formal publication.
\end{abstract}

\begin{IEEEkeywords}
codebook-based real image super-resolution, uncertainty-guided, feature matching, and align-attention.
\end{IEEEkeywords}

\IEEEpeerreviewmaketitle
%% main text

\section{Introduction}
\label{sec-intro}

Image super-resolution is a classic and active low-level vision task. It aims to reconstruct high-quality images (HQ) from low-quality images (LQ). Although deep learning has significantly improved SR methods, most existing models~\cite{edsr,rcan,rdn, niu2020single, wen2024adaptive, san,liang2021swinir,zhang2022efficient, dat} trained on simplified degradation scenarios struggle with real-world images that feature complex and unknown degradations. Thus, real-world image super-resolution (RISR) has become an increasingly popular research direction~\cite{wang2021real, xie2023desra, zhang2021designing, liang2022details}, aiming to enhance model performance in practical settings. Key challenges in RISR include constructing realistic HQ-LQ image pairs and preserving visual fidelity and clarity in recovered results. Recent approaches address the first challenges by simulating complex degradations through GAN-based methods, such as BSRGAN~\cite{zhang2021designing} and RealESRGAN~\cite{wang2021real}. Additionally, perceptual~\cite{perceptual} and adversarial losses~\cite{gan} have been introduced to address the limitations of traditional pixel-based loss functions, improving texture and detail preservation. Despite advances, GAN-based real-world SR methods often encounter issues like training instability and generation artifacts.

\begin{figure}[t]\footnotesize
	\begin{center}
		\tabcolsep 1.5pt
		\includegraphics[width=0.98\linewidth]{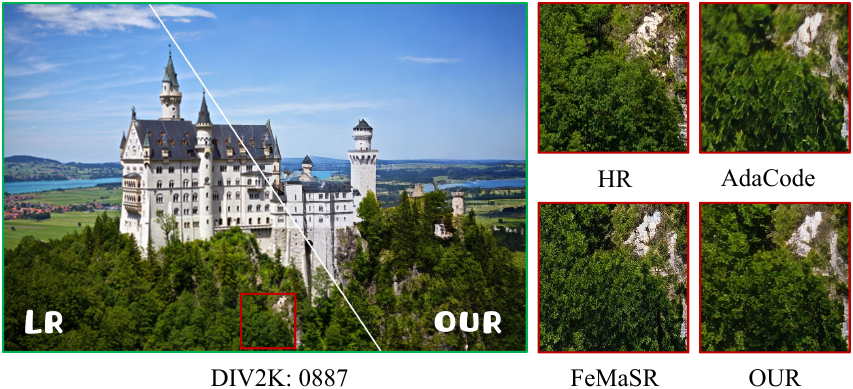}
	\end{center}
	\caption{Qualitative comparison of three VQVAE-based methods on DIV2K dataset. Our method achieves a strong balance between detail preservation and visual quality. It avoids the smearing artifacts of AdaCode~\cite{liu2023learning} and the over-sharpening of FeMaSR~\cite{chen2022femasr}. Our result is closer to the HR image, with superior fidelity and visual performance.}
	\label{fig:teaser}
\end{figure}

\begin{figure*}[t]
	\begin{center}
		\includegraphics[width=0.98\linewidth]{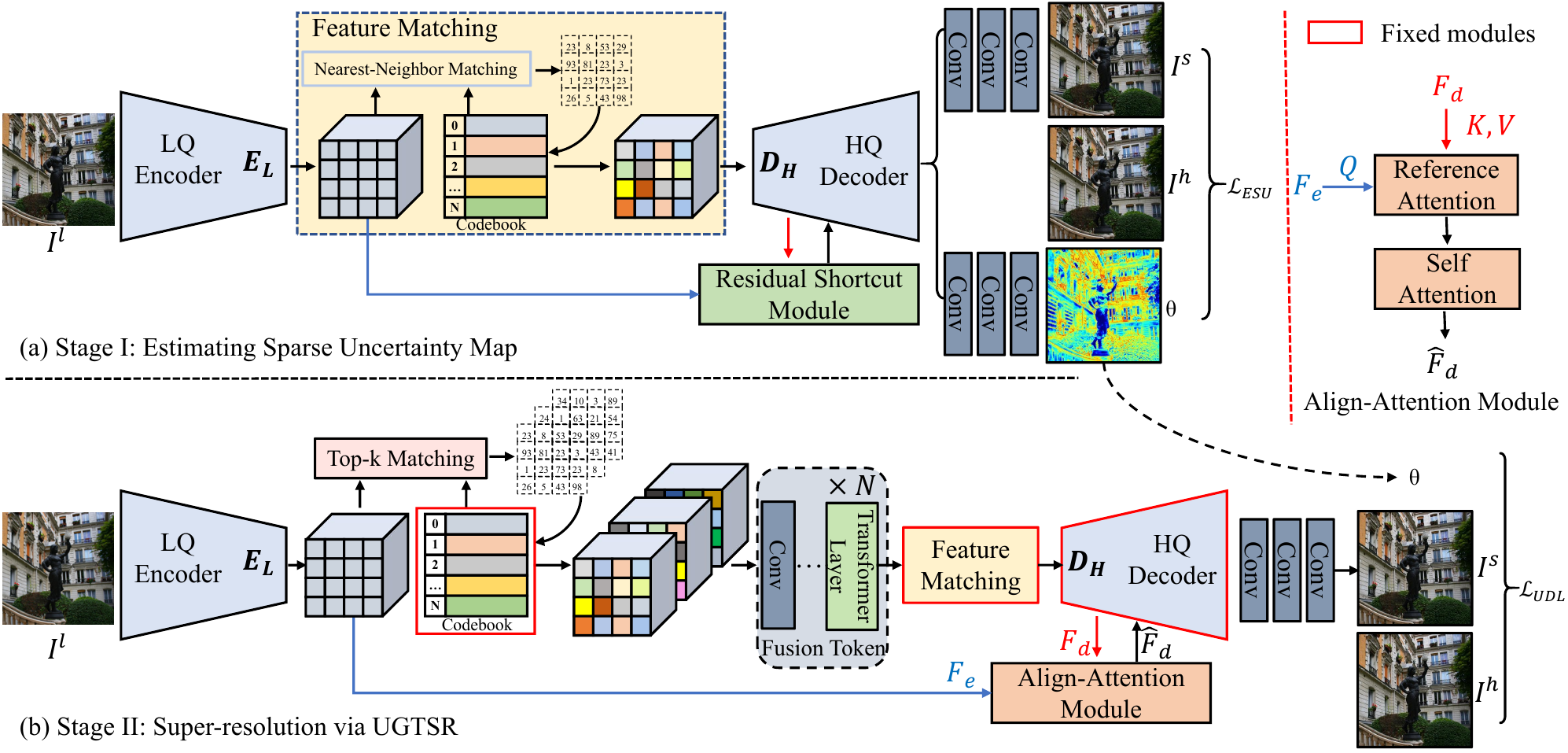}
	\end{center}
	\caption{This paper presents UGTSR, a novel uncertainty-guided two-stage super-resolution framework. In the first stage, the model leverages a pre-trained FeMaSR codebook and $\mathcal{L}_{ESU}$ loss function to estimate uncertainty distributions across image regions. Building upon this foundation, the second stage integrates an uncertainty-constrained $\mathcal{L}_{UDL}$ loss, a Top-k feature matching strategy, and an Align-Attention module to achieve high-fidelity texture reconstruction.}
	\label{fig: architecture}
\end{figure*}

The recent rise of diffusion models~\cite{ho2020denoising} has led to significant advancements across various domains, such as image generation and restoration, text-to-image synthesis, and image super-resolution. Large-scale pre-trained text-to-image (T2I) diffusion models~\cite{rombach2022high}, trained on datasets exceeding five billion image-text pairs, possess powerful natural image priors. Consequently, some approaches~\cite{wang2023exploiting, lin2023diffbir, yang2023pixel, wu2024seesr} leverage the strong generative capabilities of pre-trained T2I models to address RISR challenges. Although these diffusion-based methods can generate highly realistic image details, they typically require numerous diffusion steps, resulting in long inference times. Additionally, the presence of random Gaussian noise often compromises the fidelity and accuracy of the restored images.

With the introduction of VQVAE~\cite{vqvae} and VQGAN~\cite{esser2021taming}, a distinct class of models based on discrete codebooks has demonstrated exceptional performance in image restoration tasks. 
These models, trained extensively on high-quality images, contain robust natural image priors that effectively handle complex degradations found in real-world image restoration scenarios.
Several methods~\cite{chen2022femasr,liu2023learning,zhou2022codeformer,chen2024iter,guo2022lar,gu2022vqfr} have been proposed to address low-level vision restoration challenges by utilizing these pre-trained high-quality codebook priors. Although these codebook-based methods exhibit better computational efficiency and training stability compared to GAN-based~\cite{goodfellow2020generative} and diffusion-based~\cite{ho2020denoising} approaches, they still face several unresolved issues. We summarize these issues into three main points:

\begin{enumerate}[1)]
	\item Our observations show that although VQVAE-based methods restore sharp image details effectively, merging codebooks(e.g., AdaCode)often introduces significant painterly stylization, compromising photorealism.
	\vspace{0.5mm}
	\item Recent VQVAE-based~\cite{vqvae} methods use nearest-neighbor feature matching, but image degradation often causes inaccuracies, reducing reconstruction quality. CodeFormer~\cite{zhou2022codeformer} enhances matching accuracy with Transformer-based global modeling, but significantly increases computational complexity for high-resolution images.
	\vspace{0.5mm}
	\item Real-world SR methods based on codebook matching often suffer reduced fidelity due to misalignment between LQ and HQ features caused by image degradation.
\end{enumerate}

To address the aforementioned issues, we propose an efficient network architecture that integrates uncertainty-guided texture generation, the Top-k feature matching strategy, and an align-attention module termed UGTSR. This architecture aims to solve the challenges faced by existing methods in detail realism, feature matching accuracy, and fidelity. Our contributions can be summarized as follows:

\begin{enumerate}[1)]
	\item \textbf{Uncertainty learning to enhance realism.} We utilize the high-quality image codebook from FeMaSR~\cite{chen2022femasr} as priors for subsequent image processing. Our training framework adopts a two-stage strategy: First, we introduce an uncertainty-guided mechanism that identifies critical regions in images, generating informative uncertainty maps. Second, we assign higher weights to texture-rich regions based on uncertainty maps. Specifically, we introduce an uncertainty-driven loss function $\mathcal{L}_{UDL}$ that adaptively emphasizes high-frequency details, enhancing the fidelity and realism of the reconstructed images.
	\vspace{0.5mm}
	\item \textbf{Top-k strategy to improve accuracy in feature matching.} We observed that the conventional nearest-neighbor matching strategy used in FeMaSR~\cite{chen2022femasr} achieves only 21\% accuracy. However, when searching for the three and five features in the codebook closest to the LR feature, the probability of the correct feature appearing significantly increased to 42\% and 54\%, respectively. Based on this observation, we proposed a top-k feature matching strategy. This strategy fuses the k nearest neighbors to the LR feature, allowing the network to learn a more comprehensive and accurate representation of the LR feature, thereby improving the feature matching accuracy.
	\vspace{0.5mm}
	\item \textbf{Align-attention module to improve fidelity.} Drawing insights from \cite{gu2022vqfr}, reconstruction fidelity is closely related to the fusion strategy of LR and HR features. To overcome the limitations of direct feature concatenation employed in FeMaSR \cite{chen2022femasr}, we propose an alignment attention module, facilitating a more effective fusion of LR and HR features and thereby enhancing the fidelity of reconstructed images.
\end{enumerate}

In conclusion, we present UGTSR, a novel framework for real-world image super-resolution (RISR). We evaluate the proposed method on various synthetic and real-world datasets. Extensive experiments demonstrate that our algorithm outperforms state-of-the-art RISR approaches.

\section{Related Work}
\label{sec-related}
Deep learning has become the main technology in super-resolution (SR) reconstruction. Since SRCNN~\cite{dong2015image} first showed the potential of convolutional neural networks (CNN) in SR tasks, researchers have begun to explore the upper limit of deep learning models in image super-resolution reconstruction. As research has progressed, more and more efficient algorithms have been proposed, which demonstrate excellent performance under idealized degradation conditions, such as bicubic downsampling and Gaussian noise~\cite{edsr,rcan,rdn, san,liang2021swinir,zhang2022efficient, dat}. However, these methods exhibit limited generalization when handling real-world degradation complexities.
\begin{figure*}[t]
	\begin{center}
		\includegraphics[width=0.98\linewidth]{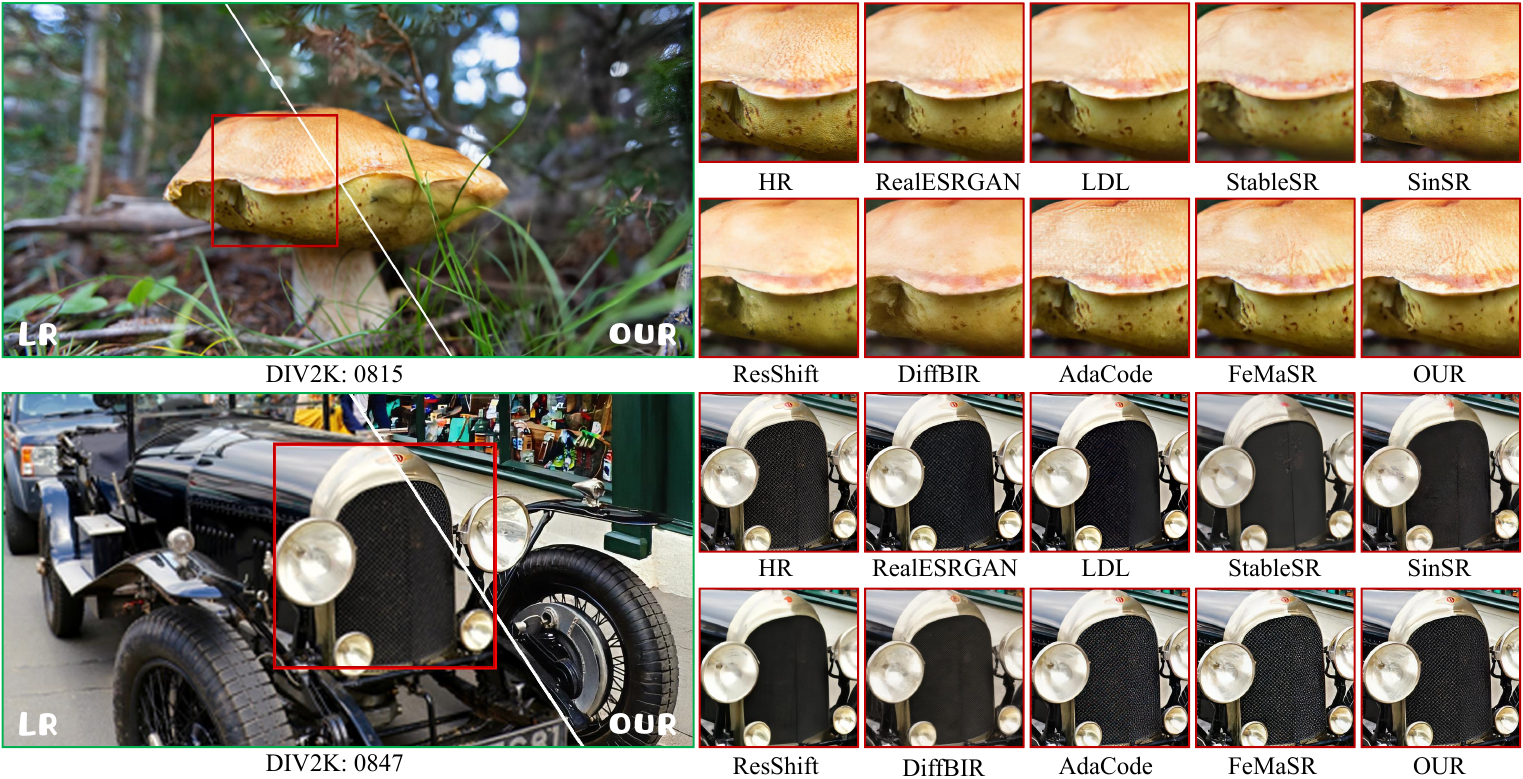}
	\end{center}
	\caption{Qualitative comparisons on DIV2K~\cite{DIV2K} dataset ($2 \times$ SR). Our method demonstrates superior artifact removal and texture restoration compared to state-of-the-art approaches. Please zoom in for the best view.}
	\label{fig: div2kx2}
\end{figure*}

\subsection{Real-World Super-Resolution}
In recent years, generative models such as Generative Adversarial Networks (GAN~\cite{goodfellow2020generative}), diffusion models~\cite{ho2020denoising}, and Vector Quantized Variational Autoencoders (VQVAE~\cite{vqvae}) have attracted widespread attention for their potential in real-world image super-resolution (RISR). 
%
% SRGAN was the first to apply GANs~\cite{goodfellow2020generative} to RISR, addressing the problem of excessive smoothing in models trained with mean squared error (MSE) loss and introducing perceptual loss, a new function combining VGG loss and GAN loss—to retain detailed textures in the restored images. 
% %
SRGAN was the first to introduce GANs~\cite{goodfellow2020generative} to RISR, effectively addressing the issue of excessive smoothing associated with models trained using mean squared error (MSE) loss. It proposed a perceptual loss function~\cite{perceptual} which can better preserve detailed textures in the reconstructed images. Subsequent studies, including BSRGAN~\cite{zhang2021designing} and RealESRGAN~\cite{wang2021real}, improved RISR performance by simulating real-world image degradation, using methods like randomly shuffled degradation and combinations of higher-order degradations. These advancements have laid a solid foundation for further RISR research, developing various GAN-based RISR methods. Although GANs~\cite{goodfellow2020generative} have achieved remarkable progress in the RISR domain, their inherent training instability and discriminator limitations with large datasets often produce unnatural artifacts. To mitigate this, research efforts like LDL~\cite{liang2022details} and DeSRA~\cite{xie2023desra} have introduced strategies to reduce artifacts. While these approaches are effective in minimizing artifacts, they often come at the cost of suppressing the generation of natural and realistic texture details, thereby limiting the overall visual quality of the results.
\begin{figure*}[t]
	\begin{center}
		\includegraphics[width=0.98\linewidth]{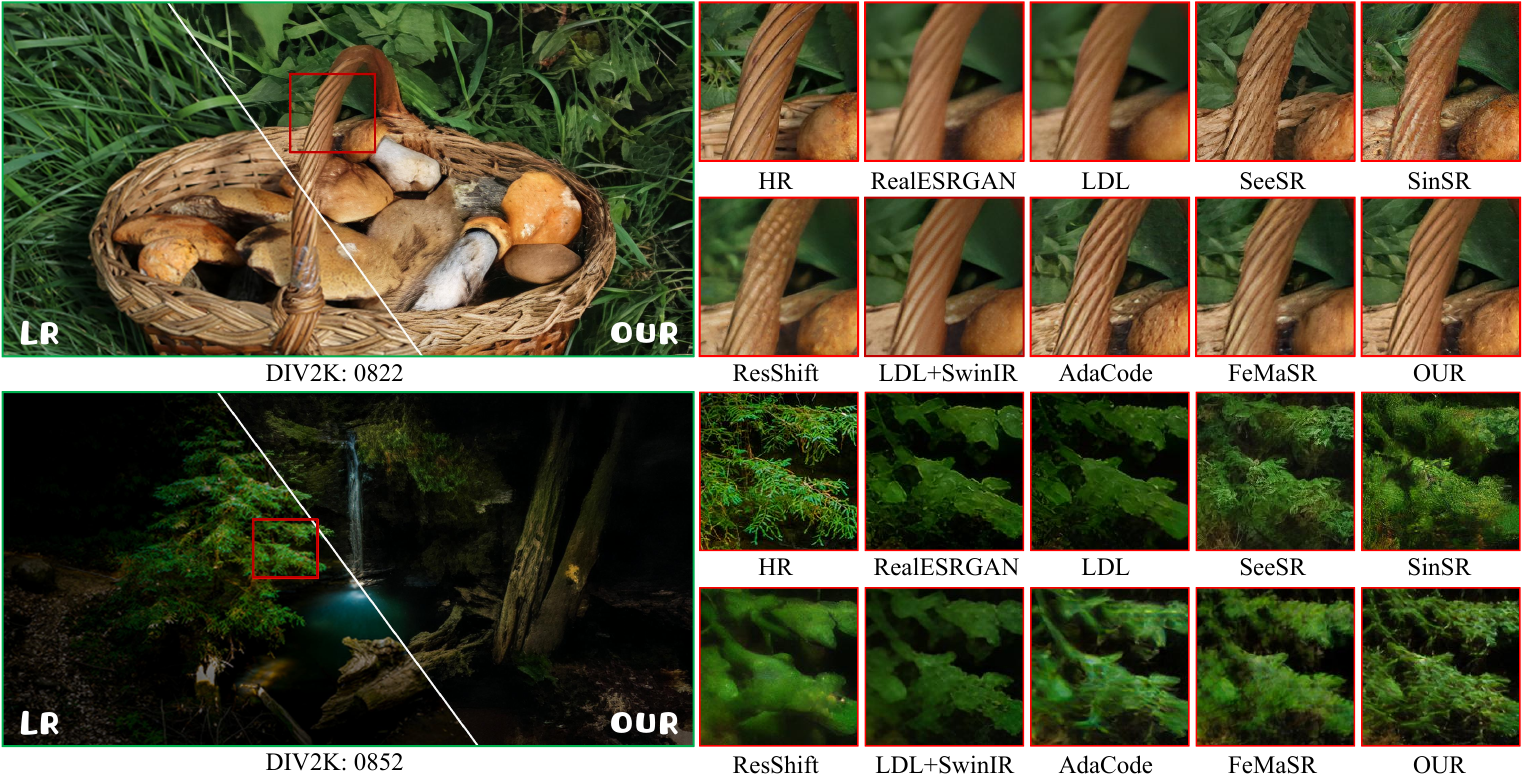}
	\end{center}
	\caption{Qualitative comparisons on DIV2K~\cite{DIV2K} dataset ($4 \times$ SR). Our method demonstrates superior artifact removal and texture restoration compared to state-of-the-art approaches. Please zoom in for the best view.}
	\label{fig: div2kx4}
\end{figure*}

With the rise of diffusion~\cite{ho2020denoising} models, some studies have attempted to apply Denoising Diffusion Probabilistic Models (DDPM)~\cite{ho2020denoising} to image SR tasks under simple degradation assumptions~\cite{kawar2022denoising, saharia2022image}. 
% However, these approaches fall short when confronted with the complexities of real-SR scenarios. 
Recently, several studies~\cite{lin2023diffbir, wang2023exploiting, wu2024seesr, yang2023pixel} have begun leveraging large-scale pre-trained text-to-image (T2I) models, such as the Stable Diffusion model (SD), to address RISR challenges. For instance, StableSR~\cite{wang2023exploiting} balances fidelity and perceptual quality by introducing a fine-tuned time-aware encoder and a feature-warping module. DiffBIR~\cite{lin2023diffbir} employs a two-stage strategy: first performing preliminary image reconstruction, followed by enhancing image details using the SD model. PASD~\cite{yang2023pixel} designed a pre-trained SD model with pixel-aware cross-attention, using low-level and high-level features extracted from low-quality images as input. SeeSR~\cite{wu2024seesr} improves semantic awareness by incorporating text and soft prompts to guide the diffusion process. Although these methods can generate more realistic image textures, they commonly suffer from long inference times and insufficient fidelity, limiting their widespread application.

\subsection{Visual Representation Dictionary Learning}
VQVAE~\cite{vqvae} introduced a novel discrete representation method by combining variational autoencoders with vector quantization to learn discrete latent representations. Specifically, the vector quantization technique maps continuous latent representations to a discrete codebook space, quantizing continuous representations into discrete codes through nearest-neighbor lookup. This discrete representation method enhances the ability to capture the discrete structure of data, improving the quality and diversity of generated samples. Building on VQVAE~\cite{vqvae}, subsequent work further improved the perceptual quality of reconstructed images by refining codebook learning. VQVAE-2~\cite{vqvae2} adopted a multi-scale structure, enabling it to model both global image information (such as shape and structure) and local information (such as texture details), thereby enhancing image reconstruction quality and detail. VQGAN~\cite{esser2021taming} introduced generative adversarial learning, perceptual loss, and refined codebook learning to further improve the realism and diversity of generated images. The trained codebook can serve as a high-quality prior, boosting the performance of many low-level vision tasks, such as face restoration~\cite{zhou2022codeformer}, low-light enhancement~\cite{liu2023low}, image dehazing~\cite{wu2023ridcp}, and image super-resolution.
CodeFormer~\cite{zhou2022codeformer} applies the codebook learning strategy to face restoration tasks, proposing the use of the transformer's global modeling capabilities to predict code sequences, thereby better utilizing codebook priors. VQFR~\cite{gu2022vqfr} introduces a parallel decoder design to merge low-level input features with VQ codebook-generated features, enhancing the restoration quality of facial details. The LLIE~\cite{liu2023low} algorithm optimizes codebook learning through residual quantization and introduces a query module to reduce the discrepancy between low-light image features and normal-light codebook features, effectively constructing a low-light image enhancement algorithm. RIDCP~\cite{wu2023ridcp} proposes the HQP matching operation to bridge the gap between synthetic and real data, leveraging the learned high-quality codebook to improve real image dehazing performance. 

Codebooks provide a compact yet expressive feature representation, allowing many SR methods to leverage codebook priors for learning image textures and details. FeMaSR~\cite{chen2022femasr} introduces the codebook prior into blind image SR, fusing LQ features from the input with matched HQ codebook prior features to enhance the quality of the reconstructed image. Unlike FeMaSR~\cite{chen2022femasr}, which relies on a single codebook prior, AdaCode~\cite{liu2023learning} learns a set of base codebooks and corresponding weight maps, achieving dynamic weighting of the codebook set. However, this weighting operation results in reconstructed images that often exhibit a painterly effect with poor fidelity. RTCNet~\cite{qin2023blind} proposes a degradation-aware texture codebook module and a low-level-friendly patch-aware texture prior module, effectively enhancing the modeling capability for the correspondence between LR and HR images. ITER~\cite{chen2024iter} combines discrete diffusion generation pipelines with token evaluation and refinement, iteratively generating image texture details, thus improving RISR performance.

% This strategy not only enhances the flexibility of the codebook but also significantly improves the expressiveness of the discrete prior codebook.
\subsection{Uncertainty Learning}
\begin{figure*}[t]
	\begin{center}
		\includegraphics[width=0.98\linewidth]{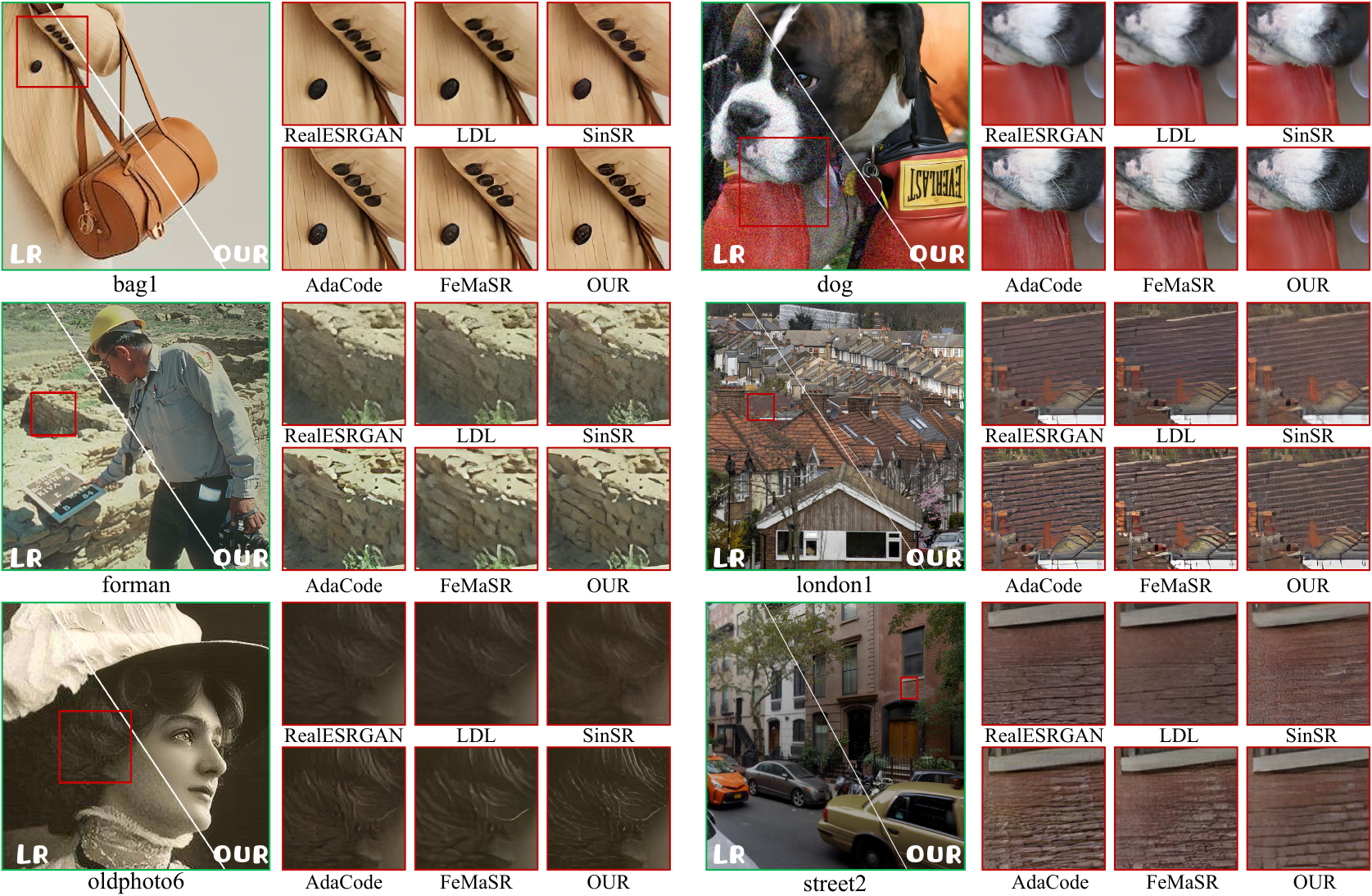}
	\end{center}
	\caption{Visual comparisons on RealSet65~\cite{yue2023resshift} dataset under $4\times$ SR. Our method achieves superior authenticity, effectively suppressing artifacts and reconstructing finer texture details compared to previous state-of-the-art approaches. Please zoom in for the best view.}
	\label{fig: realset}
\end{figure*}

Uncertainty can be categorized into two types: epistemic uncertainty, which arises from insufficient training data and reflects the model's uncertainty, and aleatoric uncertainty, which refers to the inherent noise within the observed data. The former can be mitigated by augmenting the dataset, while the latter can be captured through probabilistic classification and regression models as a result of maximum likelihood estimation. Numerous studies have introduced uncertainty into data noise regression problems~\cite{goldberg1997regression,wright1999bayesian,bishop1996regression}, thoroughly exploring its properties and corresponding treatment approaches. Recent research has revealed that modeling data uncertainty in deep learning enhances the performance and robustness of neural networks across various computer vision tasks~\cite{kendall2017uncertainties,badrinarayanan2017segnet,chang2020data}, such as image classification~\cite{gu2015active}, facial recognition~\cite{chang2020data,shi2019probabilistic}, semantic segmentation~\cite{shi2019probabilistic,lee2019gram}, pedestrian re-identification~\cite{shi2022spatial}, image super-resolution~\cite{ning2022learning,ning2021uncertainty}, and camouflaged object detection~\cite{yang2021uncertainty}.~\cite{kendall2017uncertainties}proposed a Bayesian deep learning framework that combines aleatoric and epistemic uncertainty, improving the robustness and performance of models in per-pixel semantic segmentation and depth regression tasks. \cite{chang2020data} was the first to apply data uncertainty learning to facial recognition, simultaneously learning both features (mean) and uncertainty (variance). 
The loss functions presented in the studies~\cite{kendall2017uncertainties,badrinarayanan2017segnet,chang2020data} can be concisely expressed using the following equation.
\begin{equation}
\mathcal{L} = \frac{1}{N} \sum_{i=1}^{N} \frac{|x_i - f(y_i)|_2^2}{2\sigma_i^2} + \frac{1}{2} \ln \sigma_i^2
\end{equation}
where $f(y_i)$ and $ \sigma_i^2$ represent the learned mean (image features) and variance (the uncertainty), respectively. The aforementioned loss function indeed enhances the model's robustness to data noise. In these tasks, high-uncertainty pixels are regarded as less reliable, and the loss function assigns them lower importance. 

However, contrary to the above approaches, the work by~\cite{ning2021uncertainty} proposed an uncertainty-driven loss function that focuses on high-uncertainty pixels to improve the texture details of reconstructed images in SR tasks. It was observed that pixels with high uncertainty typically belong to complex texture or edge regions, which are more critical than smooth areas. Therefore, the uncertainty-based loss function was introduced to increase focus on these high-uncertainty pixels. Building on~\cite{ning2021uncertainty},~\cite{ning2022learning} improved the visual quality of RISR by learning the degradation uncertainty of real-world LR images, and constructing paired data from multiple sampled low-resolution images to train the SR network. 

\begin{table*}[t]
    \caption{Quantitative comparison with state-of-the-art methods on synthetic benchmarks. LR images are generated with a mixed degradation model of BSRGAN~\cite{zhang2021designing} and RealESRGAN~\cite{wang2021real}. PSNR/SSIM $\uparrow$: the higher, the better; LPIPS $\downarrow$: the lower, the better. LPIPS scores can better reflect texture quality, and the best and second performances are marked in \redb{red} and \ublue{blue}.} \label{tab:synthetic}
    \setlength{\tabcolsep}{1pt}
    \renewcommand*{\arraystretch}{1.2}
    \resizebox{\textwidth}{!}{
    \begin{tabular}{|c|c|cc|c|cc|c|cc|c|cc|c|cc|c|}
    \hline
    \multirow{2}{*}{Method} & \multirow{2}{*}{Scale}  & \multicolumn{3}{c|}{Set5} & \multicolumn{3}{c|}{Set14} & \multicolumn{3}{c|}{BSD100} & \multicolumn{3}{c|}{Urban100} & \multicolumn{3}{c|}{Manga109} \\ \cline{3-17} 
     &  & PSNR & SSIM & LPIPS & PSNR & SSIM & LPIPS & PSNR & SSIM & LPIPS & PSNR & SSIM & LPIPS & PSNR & SSIM & LPIPS \\ \hline
    DiffBIR~\cite{lin2023diffbir} & $\times2$  & 24.637 & 0.7096 & 0.2215 & 23.006 & 0.5606 & 0.2990 & 23.248 & 0.5328 & \ublue{0.3543} & 21.64 & 0.5850 & 0.3081 & 22.584 & 0.7046 & 0.2523 \\
    SeeSR~\cite{wu2024seesr} & $\times2$    & 25.232 & 0.7311 & 0.2017 & 22.971 & 0.5847 & 0.2862 & 23.022 & 0.5466 & \redb{0.3482} & 21.359 & 0.6083 & \redb{0.2610} & 22.132 & 0.7416 & 0.2163 \\
    \hdashline
    RealESRGAN~\cite{wang2021real} & $\times2$  & 25.379 & 0.7598 & 0.1917 & 24.050 & 0.6344 & 0.3088 & 23.505 & 0.5740 & 0.3556 & 21.196 & 0.6193 & 0.2917 & 23.647 & 0.7812 & \ublue{0.1932} \\
    LDL+RRDB~\cite{liang2022details} & $\times2$  & 24.598 & 0.7538 & 0.2053 & 23.350 & 0.6261 & 0.3169 & 22.924 & 0.5721 & 0.3677 & 20.568 & 0.6163 & 0.3032 & 23.180 & 0.7757 & 0.2036 \\
    FeMaSR~\cite{chen2022femasr} & $\times2$  & 25.035 & 0.7375 & \ublue{0.1760} & 23.765 & 0.6068 & 0.2694 & 23.008 & 0.5356 & 0.3581 & 21.345 & 0.5949 & \ublue{0.2821} & 23.477 & 0.7456 & 0.1991 \\
    AdaCode~\cite{liu2023learning} & $\times2$ & 25.315 & 0.7305 & 0.1804 & 23.817 & 0.6020 & \ublue{0.2693} & 23.232 & 0.5454 & 0.3593 & 21.659 & 0.6063 & 0.2921 & 23.800 & 0.7525 & 0.1949 \\
    
    OUR & $\times2$  & 25.453 & 0.7512 & \redb{0.1710} & 23.997 & 0.6150 & \redb{0.2668} & 23.320 & 0.5570 & 0.3587 & 21.787 & 0.6083 & 0.2867 & 23.654 & 0.7555 & \redb{0.1870} \\

    \hline
    \hline
    DiffBIR~\cite{lin2023diffbir} & $\times4$  & 23.468 & 0.6769 & 0.2175 & 21.330 & 0.4938 & 0.3894 & 22.502 & 0.4941 & 0.3955 & 20.987 & 0.5498 & 0.3642 & 21.551 & 0.6625 & 0.3334  \\ 
    ResShift~\cite{yue2023resshift} & $\times4$   & 25.079 & 0.7386 & 0.2351 & 22.504 & 0.5524 & 0.3748 & 22.719 & 0.5034 & 0.3991 & 20.969 & 0.5582 & 0.3889 & 22.286 & 0.6983 & 0.3242 \\ 
    SeeSR~\cite{wu2024seesr} & $\times4$  & 19.136 & 0.4936 & 0.4066 & 20.191 & 0.4884 & 0.3742 & 21.601 & 0.4855 & 0.4102 & 20.870 & 0.5711 & \redb{0.3018} & 20.076 & 0.6668 & 0.3011 \\ 
    SinSR~\cite{wang2024sinsr} & $\times4$  & 24.013 & 0.6906 & \redb{0.2212} & 21.957 & 0.5181 & 0.3732 & 22.407 & 0.4844 & 0.4066 & 20.546 & 0.5314 & 0.3885 & 21.579 & 0.6513 & 0.3323 \\ 
    \hdashline
    RealESRGAN~\cite{wang2021real} & $\times4$ & 24.509 & 0.7339 & 0.2623 & 22.091 & 0.5600 & 0.3915 & 22.617 & 0.5297 & 0.4112 & 19.921 & 0.5574 & 0.3643 & 21.583 & 0.7074 & 0.3070 \\  
    LDL+SwinIR~\cite{liang2022details}  & $\times4$ & 24.112 & 0.7288 & 0.2538 & 21.669 & 0.5672 & 0.3864 & 22.299 & 0.5313 & 0.4027 & 18.929 & 0.5593 & 0.3638 & 21.127 & 0.7088 & 0.3049 \\
    LDL+RRDB~\cite{liang2022details}  & $\times4$ & 24.227 & 0.7299 & 0.2693 & 21.849 & 0.5585 & 0.3994 & 22.456 & 0.5316 & 0.4249 & 19.661 & 0.5568 & 0.3762 & 21.477 & 0.7054 & 0.3173 \\
    FeMaSR~\cite{chen2022femasr}  & $\times4$ & 23.063 & 0.6577 & 0.2492 & 21.303 & 0.5061 & 0.3885 & 21.557 & 0.4753 & 0.4077&19.820  &0.5228  &0.3574 & 21.050 & 0.6565 & 0.2972 \\
    AdaCode~\cite{liu2023learning}  & $\times4$  & 23.966 & 0.6860 & 0.2281 & 21.551 & 0.5153 & \redb{0.3597} & 22.155 & 0.5008 & \ublue{0.3923} & 20.295 & 0.5391 & 0.3482 & 21.585 & 0.6736 & \ublue{0.2843} \\

    OUR  & $\times4$ &24.595&0.7084&\ublue{0.2238}&21.853&0.5294&\ublue{0.3619}&22.357&0.5013&\redb{0.3850}&20.704&0.5547&\ublue{0.3398}&22.116&0.6832&\redb{0.2813} \\  \hline
    \end{tabular}
    }
\label{tab:benchmark}
\end{table*}

\begin{table*}[t]
\centering
\setlength{\belowcaptionskip}{-2mm}
\caption{A comprehensive evaluation against state-of-the-art methods on DIV2K-Valid datasets. PSNR/SSIM $\uparrow$: the higher, the better; LPIPS $\downarrow$: the lower, the better. LPIPS scores can better reflect texture quality, and the best and second performances are marked in \redb{red} and \ublue{blue}.}
\resizebox{0.95\textwidth}{!}{
\fontsize{12pt}{16pt}\selectfont
\begin{tabular}{c|c|c|c|c|c|c|c|c|c|c|c}
\hline
\multirow{2}{*}{Datasets} &
\multirow{2}{*}{Metrics} &
\multicolumn{4}{c|}{Diffusion-based Methods} &
\multicolumn{3}{c|}{GAN-based Methods} &
\multicolumn{3}{c}{VQVAE-based Methods} \\
\cline{3-12}
& & DiffBIR & ResShift & SeeSR & SinSR & RealESRGAN & LDL+SwinIR & LDL+RRDB & FeMaSR & AdaCode & OUR \\
\hline

\multirow{3}{*}{\textit{DIV2K-Valid$\times 2$}}
& PSNR $\uparrow$ & 24.902 & -/- & 24.608 & -/- & 25.193 & -/- & 24.441 & 24.695 & 25.099 & 25.380 \\
& SSIM $\uparrow$ & 0.6468 & -/- & 0.6640 & -/- & 0.7056 & -/- & 0.7032 & 0.6652 & 0.6761 & 0.6945 \\
& LPIPS $\downarrow$ & 0.3237 & -/- & 0.2931 & -/- & 0.2902 & -/- & 0.3054 & 0.2909 & \ublue{0.2868} & \redb{0.2721} \\
\hline

\multirow{3}{*}{\textit{DIV2K-Valid$\times 4$}}
& PSNR $\uparrow$ & 24.134 & 24.034 & 20.488 & 23.264 & 23.649 & 23.423 & 23.415 & 22.989 & 23.395 & 24.183 \\
& SSIM $\uparrow$ & 0.6218 & 0.6337 & 0.5832 & 0.5770 & 0.6571 & 0.6652 & 0.6578 & 0.6149 & 0.6291 & 0.6454 \\
& LPIPS $\downarrow$ & 0.3539 & 0.3724 & 0.3668 & 0.4080 & 0.3571 & 0.3450 & 0.3682 & 0.3382 & \ublue{0.3302} & \redb{0.3153} \\
\hline
\end{tabular}
}
\label{tab:div2k}
\end{table*}

\section{Our Method}
\label{sec-method}
% This section provides a comprehensive description of our proposed UGTSR. Our core contribution lies in the integration of uncertainty learning, which significantly enhances the capability of codebook-based SR models to recover fine details and textures. Additionally, we introduce a Top-k feature matching strategy to improve the accuracy of feature matching and to facilitate the matching of LQ features with HQ features. Furthermore, to enhance the fidelity of reconstructed images, we propose an align-attention mechanism to fuse low-resolution features with their corresponding high-resolution counterparts from the codebook.

This section provides a comprehensive description of our proposed UGTSR, which integrates uncertainty-guided learning, a Top-3 feature matching strategy, and a cross-attention mechanism. As illustrated in Figure~\ref{fig: architecture}, our core contribution lies in utilizing uncertainty to guide the generation of fine details and textures. Additionally, we introduce a Top-k feature matching strategy to improve the accuracy of feature matching and to facilitate the matching of LQ features with HQ features. Furthermore, to enhance the fidelity of reconstructed images, we propose an align-attention mechanism to fuse low-resolution features with their corresponding high-resolution counterparts from the codebook.

\subsection{Uncertainty-Guided Texture Generation}
We first evaluated the reconstruction performance of different codebook-based methods. As shown in Figure \ref{fig:teaser}, the FeMaSR~\cite{chen2022femasr} method tends to produce artifacts when handling regions with rich textural details. In contrast, the AdaCode~\cite{liu2023learning} method, due to the weighted averaging of various codebooks, generates images with a painting-like effect. Based on these observations, we realized that codebook-based methods require greater precision and refinement when dealing with highly textured regions. Building on the findings of~\cite{ning2021uncertainty}, we observed that uncertainty maps can effectively and accurately identify high-frequency regions or areas with complex texture patterns in images. Therefore, we propose using uncertainty learning in our RISR method, UGTSR, which leverages an uncertainty map to guide the generation of rich textures. This allows the model to capture and reconstruct fine-grained details more effectively in complex regions.

% Specifically, we adopt FeMaSR~\cite{chen2022femasr} as the baseline network, leveraging its pretrained codebook, which serves as high-resolution priors (HRPs) to provide rich structural information. These HRPs enable the extraction of high-quality information tokens that capture essential structural and texture details for our first stage training process. In this stage, the network training is guided by Equation~\ref{equa:esu}, which defines the objective for generating an uncertainty estimation map,
Specifically, we adopt FeMaSR~\cite{chen2022femasr} as the baseline, using its trained codebook as high-resolution priors (HRPs) to extract tokens capturing structural and texture details. The first training stage, guided by Equation~\ref{equa:esu}, focuses on generating an uncertainty estimation map,
\begin{equation}
\mathcal{L}_{ESU} = \frac{1}{N} \sum_{i=1}^{N} \exp(-s_i) \| x_i - f(y_i) \|_1 + 2s_i
\label{equa:esu}
\end{equation}
 % can be formulated as.
Here, $s_i = \ln{\theta_i}$, where $f(y_i)$ represents the SR image (mean), ${\theta_i}$ denotes the uncertainty (variance), and $x_i$ is the HR image.

As shown in Figure~\ref{fig: architecture}, we add two branches to the FeMaSR~\cite{chen2022femasr} model: one estimates the SR image (mean) and the other the uncertainty (variance). The ESU loss $\mathcal{L}_{ESU}$ is inspired by techniques used in high-level visual tasks, such as image classification~\cite{gu2015active}, and operates by disregarding pixels in regions with high uncertainty. As a result, the images generated in this stage tend to have lower sharpness, particularly in the rich texture regions, where details may be blurred or lost. 

% \redb{After this stage, we obtain an uncertainty estimation map, as shown in Figure~\cite{}.}

In the second stage of training, the uncertainty information is utilized to guide the generation of intricate textures. Following~\cite{ning2021uncertainty}, we incorporate the UD loss, which has been adapted to provide more robust supervision for texture detail generation. The formulation of the UD loss is expressed as follows: 
\begin{equation}
\mathcal{L}_{UDL} = \frac{1}{N} \sum_{i=1}^{N} \hat{s}_i \|x_i - f(y_i)\|_1
\label{equa:udl}
\end{equation}
where \( \hat{s}_i = s_i - \min(s_i) \), which represents a non-negative linear function.
The proposed uncertainty-guided mechanisms dynamically regulate feature generation through uncertainty maps, enhancing texture realism while preserving structural coherence. This novel approach demonstrates superior performance in fine-grained texture generation compared to existing methods.

\subsection{ Top-k Feature Matching Strategy}
To achieve accurate and efficient codebook feature matching, we propose the Top-k feature matching strategy. First, we provide a brief overview of VQGAN~\cite{vqgan}. For a given HR image $\bm{x} \in \mathbb{R}^{H\times W \times 3}$, the encoder $\bm{E}$ extracts a latent feature representation $\hat{z} = \bm{E}(\bm{x}) \in \mathbb{R}^{h\times w \times n_z}$, where $n_z$ is the dimension of latent vectors. Subsequently, a nearest-neighbor search is conducted to identify the vector $z_i$ that is closest to $\hat{z}_i \in \mathbb{R}^{n_z}$, which is formally defined as:
\begin{equation}
    z_i = \mathcal{Z}_k, \quad k = {\argmin_j} \| \hat{z}_i - \mathcal{Z}_j \|_2, \label{eq:quantize}
\end{equation}
where $\mathcal{Z} \in \mathbb{R}^{K \times n_z}$ is the codebook, $K$ is the codebook size, $i\in \{1,2,\ldots,h\times w\}$, $j \in \{1,2,\ldots,K\}$, and $z \in \mathbb{R}^{h \times w \times n_z}$.
The quantized feature map $z$ is then passed through the decoder $\bm{G}$ to reconstruct the image $\bm{x}'$:
\begin{equation}
    \bm{x}' = G(z) \approx \bm{x},
\end{equation}
 
To accurately match high-quality features from the codebook features, CodeFormer~\cite{zhou2022codeformer} redefines feature matching as a classification problem using cross-entropy loss to match low- and high-quality features. By employing the global attention mechanism of the Transformer, CodeFormer~\cite{zhou2022codeformer} effectively maps low-quality features to more precise codes. Nevertheless, despite improving matching accuracy, this approach requires significant computational resources due to the Transformer blocks. To address this problem, we propose a simple yet effective Top-k feature-matching mechanism that reduces computational complexity while enhancing feature-matching accuracy.

Our motivation stems from the observation that the nearest neighbor matching method used in FeMaSR~\cite{chen2022femasr} can be further improved by incorporating a Top-k nearest neighbor strategy. For the convenience of the following discussion, we define the accuracy of the Top-k strategy as the probability that the correct codebook is included among these Top-k candidates. We observed that the accuracy using nearest neighbor matching was about 21\% in FeMaSR~\cite{chen2022femasr}. In contrast, the accuracy was improved to 42\% by employing a Top-3 nearest neighbor strategy. Furthermore, when a Top-5 nearest neighbor strategy was applied, the accuracy reached approximately 54\%. This observation indicates that the correct codebook feature is typically found among the few codebook features most similar to the LR feature. Motivated by this insight, we hypothesize that integrating information from the Top-k nearest neighbors and learning a fused representation of these tokens could enhance matching accuracy. To validate this hypothesis, we propose a novel Top-k feature matching strategy.

Specifically, this strategy first obtains the $k$ nearest features for each current feature using the k-nearest neighbors method. These neighboring features are then aggregated through a fusion module to generate more refined feature representations. Following this, a nearest neighbor feature matching process is applied to quantize the continuous features into discrete representations within the codebook. Taking the Top-3 feature matching strategy as an example, this process can be expressed by the following equation:
\begin{gather}
    \{k_1, k_2, k_3\} = \arg\min_{\{j_1, j_2, j_3\}} \|\hat{z}_i - \mathcal{Z}_j\|_2, \\
    \hat{z}'_i = \text{Fusion}\big(\mathcal{Z}_{k_1}, \mathcal{Z}_{k_2}, \mathcal{Z}_{k_3}\big), \\
    z_i = \mathcal{Z}_p, \quad p = \arg\min_j \|\hat{z}'_i - \mathcal{Z}_j\|_2.
\end{gather}
where $\big(\mathcal{Z}_{k_1}, \mathcal{Z}_{k_2}, \mathcal{Z}_{k_3}\big)$ represents the three features in the codebook that are closest to $\hat{z}_i$,  $\text{Fusion}(\cdot)$ represents the fusion module. Notably, the fusion module employs the attention layer design from Restormer~\cite{Zamir2021Restormer}, which does not introduce notable additional burdens regarding parameters and computational cost. As indicated in Figure~\ref{fig:topk}, the feature matching accuracy is improved by approximately $3\%$ due to the application of the Top-3 strategy. 

\subsection{Align-Attention Module }
VQVAE-based super-resolution algorithms~\cite{chen2022femasr, liu2023learning} struggle to achieve high reconstruction fidelity, which, as explained in~\cite{gu2022vqfr}, may result from the inadequate alignment between LR and HR features that commonly lead to degraded texture fidelity in the generated images. To address this challenge, we propose an Align-Attention (AA) Module designed to align and fuse low-quality features with high-quality features.

Specifically, in the Align-Attention mechanism, the LQ feature serves as the Query, facilitating the extraction of relevant high-frequency details based on its foundational characteristics. The high-frequency information, retrieved via the codebook index, acts as the Key and Value, providing fine-grained contextual details to complement the LQ feature. This design enables the Query (LQ feature) to effectively capture high-frequency details through contextual associations with the Key and Value, thereby enhancing the fidelity of the reconstructed features. The detailed mathematical formulation is provided as follows:
\begin{equation}
    \bm{F}_{\text{align}} = \text{softmax}\left(\frac{(\bm{F}_{\text{LQ}}W_Q)(\bm{F}_{\text{HQ}}W_K)^\top}{\sqrt{d_k}}\right)(\bm{F}_{\text{HQ}}W_V)
\end{equation}
where $\bm{F}_{\text{LQ}}$ and $\bm{F}_{\text{HQ}}$ denote the low-quality features and high-quality information, respectively.  \(W_Q\), \(W_K\), and \(W_V\) represent the weight matrices used to map input features to the Query, Key, and Value spaces. \(d_k\) denotes the dimension of the Key, with its scaling function aimed at normalizing the dot-product operation to stabilize the gradients of the softmax function. The proposed align-attention module facilitates feature alignment between low-quality inputs and high-quality references, effectively resolving feature mismatch issues to enhance reconstructed image quality while preserving structural fidelity.

\subsection{Loss Function}

\begin{table*}[t]
\centering
\caption{Quantitative comparison with state-of-the-art methods on synthetic benchmarks. LR images are generated with a mixed degradation model of BSRGAN~\cite{zhang2021designing} and RealESRGAN~\cite{wang2021real}. PSNR/SSIM $\uparrow$: the higher, the better; LPIPS/DISTS/FID $\downarrow$: the lower, the better. LPIPS scores can better reflect texture quality, and the best and second performances are marked in \redb{red} and \ublue{blue}.}
\label{tab:Real}
\setlength{\tabcolsep}{4pt}
\footnotesize
\renewcommand*{\arraystretch}{1.2}
\resizebox{\textwidth}{!}{
\begin{tabular}{@{}c|c|ccccc|ccccc@{}}
\hline
\multirow{2}{*}{Method} & \multirow{2}{*}{Scale} & \multicolumn{5}{c|}{Canon} & \multicolumn{5}{c}{Nikon} \\ 
\cline{3-12}
 &  & PSNR $\uparrow$ & SSIM $\uparrow$ & LPIPS $\downarrow$ & DISTS $\downarrow$ & FID $\downarrow$ & PSNR $\uparrow$ & SSIM $\uparrow$ & LPIPS $\downarrow$ & DISTS $\downarrow$ & FID $\downarrow$ \\ 
\hline

SeeSR~\cite{wu2024seesr} & $\times4$ 
& 21.895 & 0.6887 & 0.2919 & 0.1757 & \redb{66.14} 
& 23.128 & 0.6874 & 0.2868 & \ublue{0.1782} & 55.29 \\

ResShift~\cite{yue2023resshift} & $\times4$ 
& 25.859 & 0.7496 & 0.3675 & 0.2295 & 70.56 
& 25.091 & 0.7001 & 0.3800 & 0.2394 & \ublue{54.06} \\

RealESRGAN~\cite{wang2021real} & $\times4$ 
& 26.063 & 0.7863 & \ublue{0.2610} & \redb{0.1607} & 76.02 
& 25.623 & \redb{0.7604} & \ublue{0.2849} & \redb{0.1762} & 67.94 \\

LDL+SwinIR~\cite{liang2022details} & $\times4$ 
& 25.753 & 0.7846 & 0.2638 & 0.1659 & 85.91 
& 25.301 &  \ublue{0.7547} & 0.2917 & 0.1839 & 70.11 \\

AdaCode~\cite{liu2023learning} & $\times4$ 
& 26.562 & 0.7788 & 0.2611 & 0.1705 & 73.51 
& 25.952 & 0.7420 & 0.2895 & 0.1821 & 58.88 \\

FeMaSR~\cite{chen2022femasr} & $\times4$ 
& 25.607 & 0.7702 & 0.2795 & 0.1883 & 74.12 
& 25.255 & 0.7376 & 0.3019 & 0.1998 & 62.12 \\

OUR & $\times4$ 
& \redb{26.665} & \redb{0.7909} & \redb{0.2472} & \ublue{0.1651} & \ublue{66.54} 
& \redb{26.024} & 0.7526 & \redb{0.2784} & 0.1810 & \redb{51.23} \\ 
\hline
\end{tabular}
}
\label{tab:con_nik}
\end{table*}

\begin{table*}[t]
        \caption{Quantitative comparison of NIQE under x2 and x4 super-resolution on RealSet65 dataset. NIQE $\downarrow$: the lower, the better. The top-performing and runner-up results are marked in \redb{red} and \ublue{blue}, respectively. }
	\centering
	\vspace{-3mm}
	\resizebox{0.95\textwidth}{!}{
	\begin{tabular}{ccccccccc}
	\hline
	Datasets & Bicubic& ResShift & SinSR & AdaCode & RealESRGAN & FeMaSR &LDL+RRDB & Our   \\ \hline
	RealSet65$\times 2$ &8.8996&-/- &-/-  & 8.2501 & \ublue{8.2198} & 8.5846 & 8.8030 & \redb{8.2160} \\
	RealSet65$\times 4$ &11.441 &14.3610 &12.8080    & \ublue{9.7473}  & 11.009 & 10.870 &11.484& \redb{9.4059} \\ \hline
	\end{tabular}
	}
	\label{tab:Realset65}
	\vspace{-5mm}
\end{table*}
This section presents the loss functions in our method. As depicted in Figure~\ref{fig: architecture}, in the stage $\text{I}$, we impose constraints using the ESU loss to estimate the uncertainty map, as shown in Equation~\ref{equa:esu}. In stage $\text{II}$, we weight the image texture details using the uncertainty map and optimize this process with the LUD loss, as formulated in Equation~\ref{equa:udl}. The LUD loss constraint can identify regions with rich textures through the uncertainty map and assign them greater attention, thereby enhancing the generation of details. Additionally, our method settings for other loss functions are aligned with those of FeMaSR~\cite{chen2022femasr}. We use $\mathcal{L}_{1}$ loss to supervise the difference between the generated images and the ground truth, while also incorporating GAN loss $\mathcal{L}_{adv}$~\cite{gan} and perceptual loss $\mathcal{L}_{per}$~\cite{perceptual} to enhance the semantic information of the generated images. The mathematical formulations of these loss functions are presented as follows,
\begin{gather}
\mathcal{L}_{1} = \|I_{hr} - I_{sr}\|_1 \label{eq:l1} \\
\mathcal{L}_{per} = \|\Phi(I_{hr}) - \Phi(I_{sr})\|_2^2 \label{eq:lper} \\
\mathcal{L}_{adv} = \mathbb{E}[\log D(I_{hr}) + \log(1-D(I_{sr}))] \label{eq:ladv}
\end{gather}
where $\Phi$ stands for the feature extractor of VGG-19 network. $I_{hr}$ and $I_{sr}$ denote the ground truth and super-resolved images, respectively. Additionally, we utilize $\mathcal{L}_{codebook}$ to constrain the LQ feature $\hat{z}^l$ to approach the quantized feature $z_{gt}$ from the codebook, which eases the difficulty of the feature matching process. Initially, we obtain the ground truth latent representation $z_{gt}=\mathbf{q}[E(\bm{I_{hr}}), \mathcal{Z}]$, and subsequently compute the $L_2$ loss and the Gram matrix loss for LR features:
\begin{equation}
    \mathcal{L}_{codebook} = \beta \| \hat{z}^l - z_{gt}\|_2^2 + \alpha \| \psi({\hat{z}^l}) - \psi(z_{gt}) \|_2^2,
\end{equation}  
where $\psi $ is utilized to calculate the Gram matrix of features. $\alpha$ and $\beta$ represent Gram matrix loss weight and $L_2$ loss weight, respectively. By combining multiple loss functions, our model can generate richer texture details while effectively suppressing artifacts. The total loss functions for the different stages are defined as follows.

\noindent \textbf{Stage I: Estimating Sparse Uncertainty Map}
\begin{equation}
    \mathcal{L}_{stage1} = \mathcal{L}_{codebook} + \mathcal{L}_{1} +  \mathcal{L}_{per} +\lambda_{adv}\mathcal{L}_{adv} + \mathcal{L}_{ESU}
\end{equation}

\noindent \textbf{Stage II: Super-resolution via UGTSR}
\begin{equation}
    \mathcal{L}_{stage2} = \mathcal{L}_{codebook} + \mathcal{L}_{1} +  \mathcal{L}_{per} +\lambda_{adv}\mathcal{L}_{adv} + \mathcal{L}_{UDL}
\end{equation}
where the weights for each loss are set as: $\alpha = 1, \beta = 0.25, \lambda_{adv}=0.1$.

\section{Experiments}
\label{sec-exper}
This section first presents the datasets and implementation details. Subsequently, we evaluate image restoration performance through comprehensive qualitative and quantitative comparisons. Finally, we perform ablation studies to evaluate the effectiveness of the proposed modules.

\subsection{Experimental Settings}
\noindent \textbf{Training Datasets.} Following~\cite{chen2022femasr, zhang2021designing}, we develop a trining set that includes DIV2K~\cite{DIV2K}, Flickr2K~\cite{flick2k}, DIV8K~\cite{gu2019div8k}, and 10,000 facial images from FFHQ~\cite{ffhq}.
The training data generation process is as follows: we first cropped the data into image patches of size $512 \times 512$. For the selected 10,000 aligned facial images, we resized them within the range of $[0.5, 1.0]$, followed by cropping the facial regions. For specific details, please refer to the FeMaSR~\cite{chen2022femasr} method. Finally, we obtained 198,061 high-quality $512 \times 512$ image patches. We employ BSRGAN~\cite{zhang2021designing} degradation to generate corresponding LR images. We tested on synthetic degraded and real-world image sets to provide a more comprehensive comparison of model performance.

\noindent \textbf{Synthetic Testing Datasets.} For the synthetic test set, we applied the BSRGAN~\cite{zhang2021designing} and RealESRGAN~\cite{wang2021real} mixed degradation model to degrade images from DIV2K Valid~\cite{DIV2K} and five benchmark datasets, i.e., Set5~\cite{bevilacqua2012low}, Set14~\cite{zeyde2010single}, BSD100~\cite{martin2001database}, Urban100~\cite{huang2015single}, and Manga109~\cite{matsui2017sketch}. This diverse test set provides a broader basis for model evaluation.

\noindent \textbf{Real-world Testing Datasets.} Additionally, we evaluated the SR reconstruction results on real-world datasets, comparing the performance of various methods. The real datasets include Canon~\cite{wang2021towards}, Nikon~\cite{wang2021towards}, and RealSet65~\cite{yue2023resshift}. Canon and Nikon datasets were captured using fixed digital single-lens reflex (DSLR) cameras with different focal lengths, each consisting of 50 images. In contrast, the RealSet65~\cite{yue2023resshift} is collected from the internet and exhibits more complex and severe degradation compared to the other datasets.

\subsection{Evaluation Metrics.}
To comprehensively evaluate the performance of various methods, we employed a series of both full-reference and no-reference evaluation metrics. The full-reference metrics include PSNR and SSIM~\cite{ssim2004}, calculated on the Y channel in the YCbCr color space. Additionally, we utilized LPIPS\cite{zhang2018perceptual} and DISTS~\cite{dists} as reference-based perceptual quality indicators to assess the human perceptual quality of different methods. To evaluate real-world datasets, we employed the commonly used no-reference metrics FID~\cite{heusel2017gans} and NIQE~\cite{2012niqe}.

\subsection{Implementation Details.}
% During the pre-training phase, we obtained the high-resolution priors codebook from FeMaSR~\cite{chen2022femasr}, which was subsequently utilized for uncertainty map prediction and SR training. 
In the training process, we employed the Adam optimizer~\cite{kingma2014adam}, with hyperparameters set to $\beta_1$ = 0.9 and $\beta_2$ = 0.99, while both the generator and discriminator learning rates were fixed at 0.0001. For predicting the uncertainty map during the first stage, the codebook $Z$ and decoder were kept constant, and the Top-k strategy and AA module were not applied. During stage II, the codebook $Z$, decoder, and uncertainty map prediction model remain fixed. The network was trained with a batch size of 32 and the HR image patch size was fixed to 256 $\times$ 256 for $2 \times$ and $4 \times$ SR tasks. Our algorithm was implemented using the PyTorch deep learning framework. The first stage for predicting the uncertainty map took approximately two days on four GeForce RTX 3090 GPUs, while the second training stage required about four days on the same hardware.

\begin{figure*}[t]
	\begin{center}
		\includegraphics[width=0.98\linewidth]{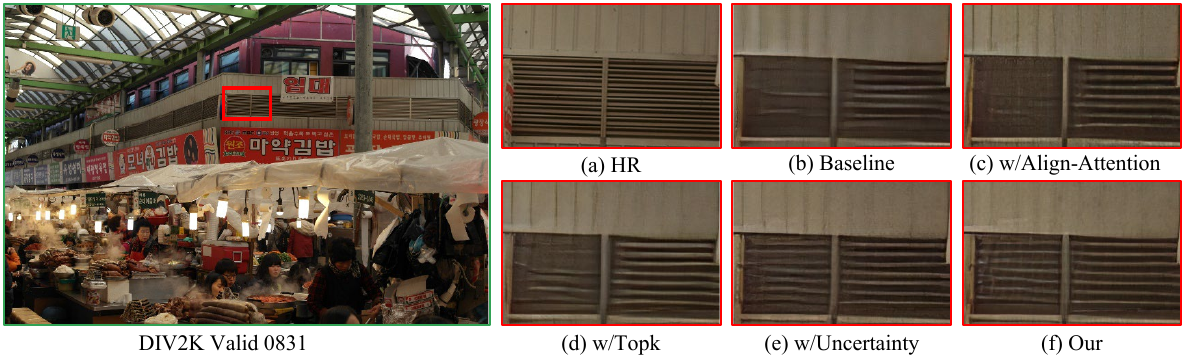}
	\end{center}
	\caption{Ablation studies for 4$\times$ SR with different model variations on the DIV2K~\cite{DIV2K} dataset. Our method UGTSR with all the proposed modules outperforms other configurations.}
	\label{fig:ablation}
\end{figure*}
\subsection{Comparison with Existing Methods.}
We conduct a comparative analysis of the proposed UGTSR method against three categories of mainstream RISR methods. The first category consists of codebook-based methods that are closest to our approach, such as AdaCode~\cite{liu2023learning} and FeMaSR~\cite{chen2022femasr}. The second category includes GAN-based methods, which comprise LDL+RRDB~\cite{liang2022details}, LDL+SwinIR~\cite{liang2022details}, and RealESRGAN~\cite{wang2021real}. The third category is the most advanced real image SR methods based on diffusion models (DM-based), including ResShift~\cite{yue2023resshift}, DiffBIR~\cite{lin2023diffbir}, SeeSR~\cite{wu2024seesr}, and SinSR~\cite{wang2024sinsr}. These three categories represent the most influential techniques for RISR reconstruction in current research. We utilized the official code and pre-trained models to obtain experimental results for most methods.

\noindent \textbf{Comparison on Synthetic Datasets.} Our experiments conducted on five benchmark synthetic datasets employed comparative performance analyses. Within the RISR domain, researchers prioritize the LPIPS metric owing to its heightened perceptual alignment with human visual assessments.
As shown in Table~\ref{tab:benchmark}, among three prevalent RISR frameworks, our method achieved the best overall performance on all five datasets for both $2 \times$ and $4 \times$ super-resolution tasks. Given that these benchmark datasets typically contain limited data volumes or lower image resolutions, we conducted additional experiments on DIV2K~\cite{DIV2K}, as presented in Table~\ref{tab:div2k}. Our method demonstrates superior performance in LPIPS metrics, specifically outperforming AdaCode by margins of 0.0147 and 0.149 for $2 \times$ and $4 \times$ SR, respectively. 

In addition, we provide visual comparisons among different methods. As illustrated in Figure~\ref{fig: div2kx2} and Figure~\ref{fig: div2kx4}, our method achieves superior fidelity and more natural texture details compared to existing approaches. Specifically, in Figure~\ref{fig: div2kx2}, our method generates more natural and rich textures on the ``mushroom", whereas AdaCode, while capable of generating abundant textures, exhibits noticeable local smearing artifacts. Other GAN-based and Diffusion-based RISR methods tend to over-smooth the mushroom textures. Similar observations can be found in sample ``0847", where VQVAE-based methods like AdaCode and FeMaSR successfully recover grid patterns but produce over-sharpened textures. In contrast, our method maintains better fidelity and more realistic detail preservation. Similarly, in Figure~\ref{fig: div2kx4}, the textures generated by our method more closely approximate the ground truth, while other methods suffer from either painting-like artifacts or compromised texture fidelity.

\noindent \textbf{Comparison on Real-World Datasets.} For real-world datasets, we conducted comparative experiments on two types of data: the Canon and Nikon datasets, which consist of HR-LR image pairs captured by DSLR cameras, and RealSet65, a collection of degraded low-resolution images gathered from the internet.
On the Canon and Nikon datasets, we evaluated $4 \times$ super-resolution reconstruction performance. As shown in Table~\ref{tab:con_nik}, our method achieved superior results across most metrics. Notably, we achieved LPIPS improvements of 0.0138 and 0.0065 over the second-best method, RealESRGAN. 

For the RealSet65 dataset, we employed the no-reference NIQE metric for evaluation. Our method achieved state-of-the-art performance for both $2 \times$ and $4 \times$ super-resolution tasks. As shown in Table~\ref{tab:Realset65}, our method obtains a 0.3414 improvement compared to the VQVAE-based approach AdaCode in the $4 \times$ super-resolution task. Visual comparisons on RealSet65, as shown in Figure~\ref{fig: realset}, highlight the superior ability of our method to generate natural and visually appealing textures. For instance, in the ``london1" sample, our results exhibit more realistic and layered textures, while AdaCode and FeMaSR, despite generating abundant textures, produce relatively messy and unrealistic results. GAN-based methods and diffusion-based approaches tend to over-smooth textures.
\begin{figure}[t]
	\begin{center}
		\includegraphics[width=0.95\linewidth]{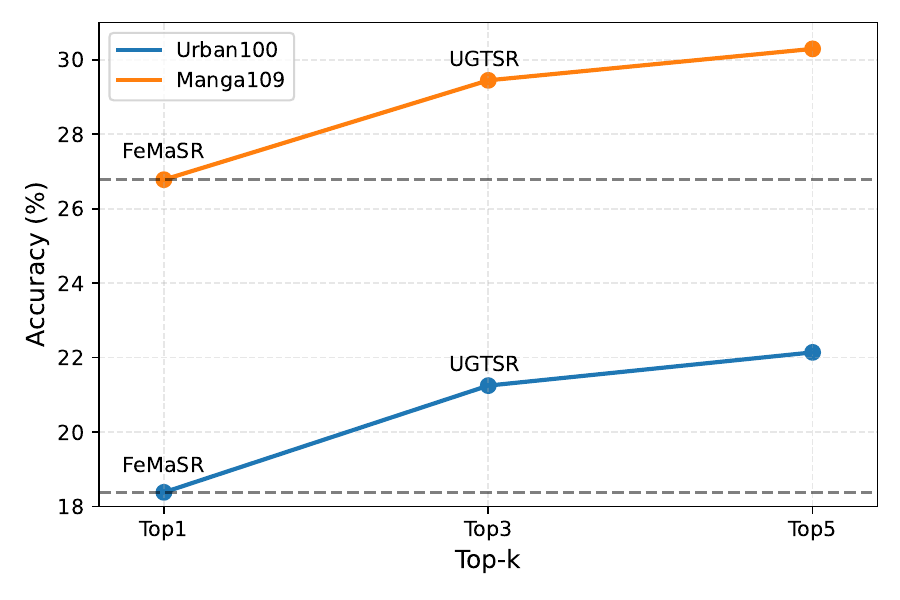}
	\end{center}
	\caption{Comparison of feature matching accuracy across different methods on the Urban100 and Manga109 datasets. We fuse Top-k information to improve feature-matching accuracy.}
	\label{fig:topk}
\end{figure}
Additionally, our method successfully preserves rich texture details while eliminating degradation. As presented in the ``oldphoto6" and ``dog" samples, our approach is capable of generating realistic hair and whisker textures. The AdaCode method produces painting-like artifacts (as seen in ``oldphoto6") or unnatural textures (such as white artifacts on the boxing gloves in ``dog"). In contrast, GAN-based and diffusion-based methods exhibit over-smoothing.
The superior performance of UGTSR stems from three key mechanisms: uncertainty-guided texture synthesis preserving fine details, precision-driven Top-k feature matching enhancing for accuracy enhancement, and Align-Attention-based feature alignment between LR features and HQ codebook features. Experimental results demonstrate that our approach achieves robust high-fidelity reconstruction on real-world datasets.

\begin{table}[t]\small
        \caption{Ablation study on DIV2K Valid dataset under $4 \times$ SR. The best and second-best results are marked in \redb{red} and \ublue{blue}, respectively.}
	\centering
	\vspace{-3mm}
	\resizebox{1.0\columnwidth}{!}{
	\begin{tabular}{ccccccc}
	\hline
	Experiments & Uncertainty & Top-k & Align-Attention & PSNR & SSIM & LPIPS   \\ \hline
	Baseline           & $\times$    & $\times$ & $\times$ & 23.592 & 0.6330 & 0.3408 \\
	w/Uncertainty           & $\checkmark$ & $\times$ & $\times$ & 23.860 & 0.6406 & \ublue{0.3261} \\
	w/Top-3           & $\times$    & $\checkmark$ & $\times$ & \ublue{24.021} & \ublue{0.6416} & 0.3391 \\
	w/Top-5           & $\times$    & $\checkmark$ & $\times$ & 23.754 & 0.6361 & 0.3338 \\
	w/Align-Attention           & $\times$    & $\times$ & $\checkmark$ & 23.296 & 0.6305 & 0.3321 \\
	OUR           & $\checkmark$ & $\checkmark$ & $\checkmark$ & \redb{24.312} & \redb{0.6474} & \redb{0.3213} \\ \hline
	\end{tabular}
	}
	\label{tab:ablation}
	\vspace{-5mm}
\end{table}

\subsection{Ablation Study}
\noindent \textbf{Effectiveness of Uncertainty Learning Strategy.} Recent studies have demonstrated the effectiveness of the uncertainty learning strategy in restoring texture details~\cite{ning2021uncertainty}. As presented in Table~\ref{tab:ablation}, we incorporated uncertainty learning into the baseline model, resulting in noticeable improvements across all evaluation metrics. Our method shows performance improvements: PSNR increased by 0.27 dB, SSIM improved by 0.076, and the LPIPS metric outperformed the baseline. Among evaluation metrics, LPIPS better reflects human visual perception, thus serving as our primary performance indicator.

As illustrated in Figure~\ref{fig:ablation}, the comparison between (b) and (e) demonstrates that incorporating uncertainty learning significantly improves the restoration of texture details in the generated results. 
Compared to (b), (e) shows richer detail, especially in the texture of the blinds. The lines in (e) are clearer, and the overall quality is closer to that of the HR image (a). The improvement is attributed to the uncertainty guidance strategy, which adaptively focuses the model on texture-rich regions, producing more realistic and visually coherent details.

\begin{figure}[t]
	\begin{center}
		\includegraphics[width=0.95\linewidth]{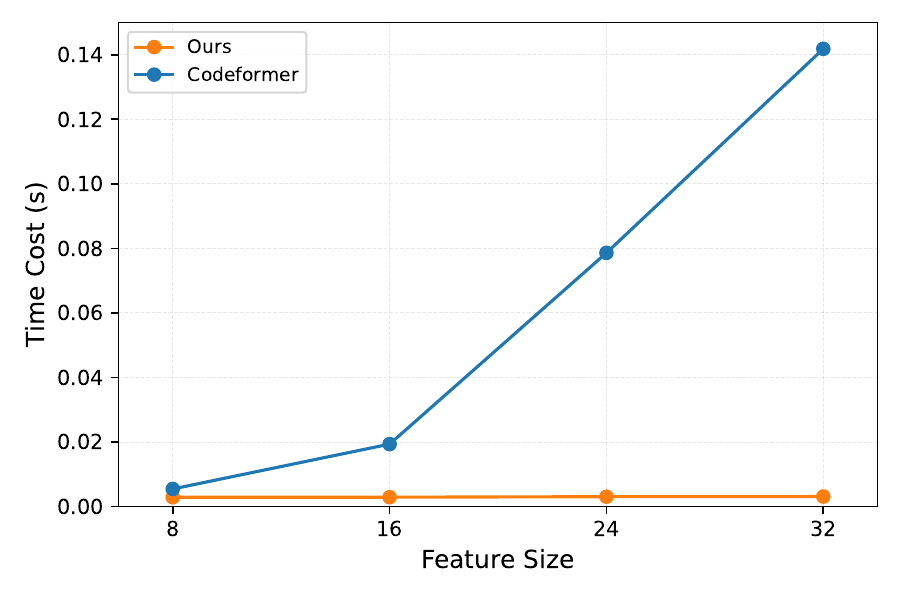}
	\end{center}
	\caption{Computational complexity comparison between our proposed Top-k strategy and the feature matching strategy in CodeFormer~\cite{zhou2022codeformer}. While CodeFormer's computational time increases quadratically with feature dimensions due to its transformer-based architecture, our Top-k strategy maintains linear time complexity.
}
	\label{fig:time}
\end{figure}

\noindent \textbf{Effectiveness of Top-k Matching.} The Top-k matching strategy significantly improves the accuracy of feature matching by about 3\%, as illustrated in Figure~\ref{fig:topk}. Its flexibility in feature selection effectively enhances matching quality. Benefiting from this improvement, the model demonstrates superior performance in both quantitative metrics and visual results. In the w/Top-k experiment, we integrated the Top-k matching module into the baseline model, and experimental results demonstrated significant performance enhancements, with PSNR increasing by 0.429 dB and SSIM improving by 0.0086, respectively. As shown in Figure~\ref{fig:ablation}, the comparison between (b) and (d) highlights that the Top-k matching strategy better restores complex edge structures. 
We also compared the results of the nearest neighbor (Top-1), Top-3, and Top-5 approaches. As indicated in Figure~\ref{fig:topk}, the results show that replacing the nearest neighbor with Top-3 and Top-5 approaches led to approximate accuracy improvements of 3\% and 4\%, respectively. This trend of accuracy enhancement was consistent across the Urban100 and Manga109 datasets, thereby validating the effectiveness of the feature matching module. The codeformer uses the Transformer module for feature prediction, resulting in an approximate 3\% improvement in accuracy compared to the nearest neighbor method. Notably, our Top-3 module demonstrated significant efficiency advantages over Codeformer. As illustrated in Figure~\ref{fig:time}, the time consumption of our method exhibits linear growth with an increase in codebook size, whereas Codeformer's method shows quadratic growth. This indicates that our approach achieves higher efficiency than Codeformer. Additionally, as presented in Table\ref{tab:ablation}, the LPIPS metric for Top-5 demonstrates a relative enhancement over Top-3, whereas PSNR and SSIM show slight decreases. Considering efficiency and performance, we ultimately adopted the Top-3 feature matching strategy in our model.

\noindent \textbf{Effectiveness of the Align-Attention Module.} VQVAE-based super-resolution methods often suffer from reduced reconstruction fidelity. According to~\cite{gu2022vqfr}, this issue may primarily arise from mismatches between LR and HR features. To address this challenge, we propose an Align-Attention (AA) module, which explicitly aligns low-quality features (Q) with high-quality features (as keys K and values V).
Experimental results show that integrating the AA module into the baseline model (w/Align-Attention) leads to a noticeable improvement of 0.0087 in LPIPS, a metric closely correlated with human visual perception. Despite slight reductions in PSNR and SSIM, our results indicate that the AA module significantly enhances the perceptual quality of reconstructed images.

\section{Conclusions}
\label{sec-con}
This paper aims to address the issues of unrealistic textures and low fidelity in codebook-based RISR methods. To mitigate the artifacts and ``painterly" effects observed in existing codebook-based SR methods, we introduce an uncertainty learning mechanism to UGTSR. This mechanism guides the network to focus on complex textured areas, enabling better control over the generated textures. To obtain more accurate HQ information during feature matching, we propose a simple yet effective Top-k matching strategy. This approach introduces a small number of additional parameters to incorporate more correct codes into the network, thereby enhancing the overall quality of the restoration. Moreover, to improve image reconstruction fidelity, we design an Align-Attention module in the decoder stage to effectively fuse low- and high-quality features. Benefiting from these designs, our method demonstrates outstanding representational capability and efficiency. Experimental results validate the superiority and effectiveness of our approach.

\ifCLASSOPTIONcaptionsoff
  \newpage
\fi

{
\bibliographystyle{IEEEtran}
\bibliography{realsr}
}

\end{document}